\documentclass[a4paper, 10 pt, conference]{ieeeconf}
\IEEEoverridecommandlockouts
\usepackage{float}
\usepackage{epsfig}
\usepackage{amsmath}
\usepackage{array}
\usepackage{amssymb}
\usepackage{cite}
\usepackage{booktabs}
\usepackage{multirow}
\usepackage{subcaption}
\usepackage{booktabs}
\usepackage{mathtools}
\usepackage{siunitx}
\usepackage{comment}
\usepackage{threeparttable}
\usepackage{bm}
\usepackage{pifont}% http://ctan.org/pkg/pifont
\makeatletter
\newcommand{\subsubsubsection}{\@startsection{paragraph}{4}{\z@}%
  {1.0\Cvs \@plus.5\Cdp \@minus.2\Cdp}%
  {.1\Cvs \@plus.3\Cdp}%
  {\reset@font\sffamily\normalsize}
}
\makeatother
\usepackage{color}
\usepackage{soul}

\usepackage{amsmath}
\usepackage{algorithmicx}
\usepackage[ruled]{algorithm}
\usepackage[noend]{algpseudocode}

\begin{document}
\title{Modular Vacuum-Based Fixturing System \\for Adaptive Disassembly Workspace Integration}

\author{Haohui Pan$^{1}$, Takuya Kiyokawa$^{1}$, Tomoki Ishikura$^{2}$, \\Shingo Hamada$^{2}$, Genichiro Matsuda$^{2}$, and Kensuke Harada$^{1}$%
\thanks{$^{1}$Department of Systems Innovation, Graduate School of Engineering Science, The University of Osaka, Toyonaka, Osaka, Japan.}%
\thanks{$^{2}$Manufacturing Innovation Division, Panasonic Holdings Corporation, 2-7 Matsuba-cho, Kadoma, Osaka, Japan.}
}

\maketitle

\begin{abstract}
The disassembly of small household appliances poses significant challenges due to their complex and curved geometries, which render traditional rigid fixtures inadequate. In this paper, we propose a modular vacuum-based fixturing system that leverages commercially available balloon-type soft grippers to conform to arbitrarily shaped surfaces and provide stable support during screw-removal tasks. To enable a reliable deployment of the system, we develop a stability-aware planning framework that samples the bottom surface of the target object, filters candidate contact points based on geometric continuity, and evaluates support configurations using convex hull-based static stability criteria. We compare the quality of object placement under different numbers and configurations of balloon hands. In addition, real-world experiments were conducted to compare the success rates of traditional rigid fixtures with our proposed system. The results demonstrate that our method consistently achieves higher success rates and superior placement stability during screw removal tasks.

Index Terms - Disassembly Automation, Soft Robotics, Vacuum-Based Fixturing, Stability-Aware Planning, Small-scale Appliances.

\end{abstract}

\IEEEpeerreviewmaketitle

\section{Introduction}

As demand for sustainable manufacturing and the circular economy grows, robotic disassembly is increasingly recognized as a key enabler for end-of-life product recovery. However, traditional disassembly systems often rely on rigid splinting and predefined trajectories, which limit adaptability to varied object geometries and uncertain contact conditions \cite{chang2017approaches, lee2001disassembly, tang2001integrated, wang2013intelligent}. These constraints are especially pronounced when dealing with delicate or irregular components, where precise contact control is required.

Soft robotics presents a promising alternative that offers morphological adaptability, safe interaction, and a compliant contact force distribution. In particular, soft robotic grippers have demonstrated effectiveness in handling fragile, deformable, or non-uniform objects \cite{crowley20223d, dontu2024applications, gafer2020quad}. Several recent designs incorporate jamming mechanisms \cite{li2017passive}, vacuum-based actuation \cite{gabriel2020modeling}, or modular balloon structures \cite{kemmotsu2024balloon} to achieve variable stiffness and passive conformity. Topology-optimized grippers \cite{pinskier2024diversity} and classification-driven approaches \cite{samadikhoshkho2019brief} have further enhanced the performance of task-specific manipulation.

Despite these advancements, the application of soft actuation principles in disassembly-oriented fixturing remains underexplored. Unlike grasping, disassembly tasks often involve asymmetric pull and press loads from screwdrivers or cutting tools. These force interactions pose unique challenges to stability and repeatability, especially in constrained or cluttered workspaces \cite{hu2023reducing, nadeau2025stable}. Although intelligent disassembly planning has been studied from a systems perspective \cite{wang2013intelligent},  the adaptability of the geometric uncertainty at the fixture level is still lacking.

To address this gap, we propose a balloon-based adaptive fixturing system composed of suction modules arranged in multipoint configurations. Inspired by the shape-memorable mechanisms of adaptive pin arrays \cite{shi2021development} and soft jig platforms \cite{kiyokawa2021soft}, our approach combines surface completeness evaluation with convex hull-based stability analysis \cite{meeran1997optimum}. A novel scoring mechanism incorporates local curvature, internal hollowness, and suction feasibility to select stable support regions \cite{funahashi1996grasp, yamada2005grasp, pozzi2016grasp}. The resulting fixture plan ensures robust placement while tolerating non-ideal alignments and tool-induced perturbations.

To evaluate the effectiveness of our proposed method, we conduct a comparative simulation between two-point and three-point suction configurations. We also compared the performance between the rigid gripper and our system in real-world disassembly tasks. Our analysis focuses on stability under press loads, repeatability of screw removal, and suction feasibility, thus demonstrating the advantages of stability-aware planning in practical scenarios.

This work contributes to bridging the fields of soft robotic manipulation and fixturing for automated disassembly. It enables geometry-aware suction planning and promotes stable object support without relying on custom rigid holders or precision placement.

\section{Related Work}

\subsection{Mechanically Adaptive Fixtures}

Mechanically adaptive fixtures dynamically conform to object geometries using physical mechanisms. Shape-memorable pin arrays \cite{shi2021development}, passive granular jamming \cite{li2017passive}, and soft-jig structures \cite{kiyokawa2021soft} have been developed to offer both compliance and support stiffness. Such methods are highly effective for objects with irregular contours, especially in collaborative or fragile part handling. However, their application has been predominantly in assembly scenarios. For disassembly, challenges arise due to asymmetric loading and fixture fatigue under repeated tool interactions \cite{chang2017approaches, lee2001disassembly}.

Some works also explore the optimization of fixture layout for curved objects \cite{ding2002fixture}, but typically from a rigid CAD-driven fixture perspective. Our work expands this scope by introducing soft, suction-based contact planning for dynamic object support.

\subsection{Soft Grippers and Suction Mechanisms}

Soft robotic grippers exploit material compliance, pneumatic actuation, and distributed contact to achieve adaptive manipulation \cite{crowley20223d, dontu2024applications, gafer2020quad}. Classification efforts highlight the diversity of gripper architectures tailored for specific tasks \cite{samadikhoshkho2019brief}. Among them, vacuum-based systems are popular for their passive conformity and active control of the gripping force \cite{gabriel2020modeling, kemmotsu2024balloon}.

Recent studies have introduced modular jamming actuators \cite{li2017passive}, topology-optimized structures \cite{pinskier2024diversity}, and soft-rigid hybrids \cite{gafer2020quad}. These innovations facilitate object handling across a wide shape spectrum, while enabling safe human-robot interaction. Yet, few of these systems are evaluated under the quasi-static and dynamic conditions involved in disassembly operations.

\subsection{Stability and Placement Planning}

Grasp stability and robustness to placement are critical for ensuring that objects remain stationary under external perturbations. Classical quality metrics focus on contact wrench space or force closure under frictional constraints \cite{liu2004quality, funahashi1996grasp}. Later work incorporates curvature-based contact reasoning \cite{yamada2005grasp}, force distribution balance \cite{pozzi2016grasp}, and grasp robustness under actuation limits \cite{hu2023reducing}.

Recent planning frameworks also consider uncertainties in object geometry and pose, which are particularly relevant for unstructured disassembly environments \cite{nadeau2025stable}. Optimization algorithms utilizing convex hull analysis \cite{meeran1997optimum} and local search \cite{tang2001integrated} offer tractable solutions in such scenarios.

Our method builds on these foundations by integrating geometric filtering, suction modeling, and contact force evaluation into a unified stability-aware fixturing pipeline tailored for automated screwing tasks.

\section{Proposed Method}

\subsection{Assumptions}

To support stability analysis and contact planning, several assumptions are made regarding fixture mechanics, object geometry, and disassembly setup.

First, suction forces from balloon-based end-effectors (CONVUM, SGB-10\footnote{https://convum.co.jp/products/en/other-en/sgb/}) are assumed to act primarily in the vertical direction. This reflects the working principle of vacuum-driven soft actuators. Preliminary tests confirm that typical tool weight (e.g., Panasonic screwdriver) is sufficient to fully compress the balloons, justifying the omission of detailed deformation modeling under quasi-static conditions.

All objects are represented by CAD-based STL models with non-planar bottom surfaces. It is assumed that these surfaces are accessible and suitable for point sampling. The system accommodates irregular geometries through contact planning and localized vertical support.
\begin{figure}[t]
    \centering
    \includegraphics[width=1\linewidth]{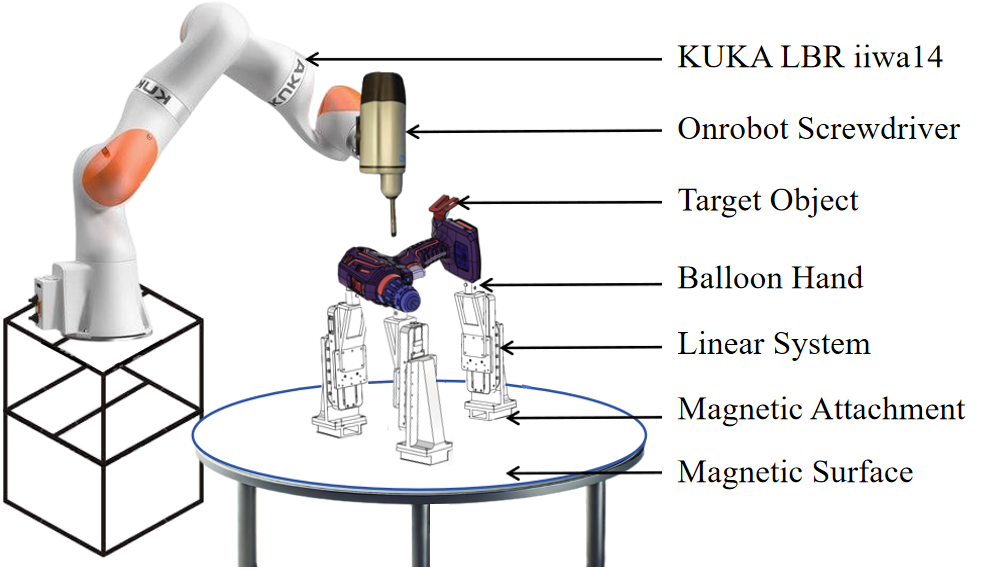}
    \caption{Overview of the robotic disassembly system consisting of a KUKA LBR iiwa14 arm, Onrobot screwdriver, balloon-hand-based support system, and magnetic attachments for fixture stabilization.}
    \label{fig:o}
\end{figure}
\begin{figure*}[t]
    \centering
    \includegraphics[width=1\linewidth]{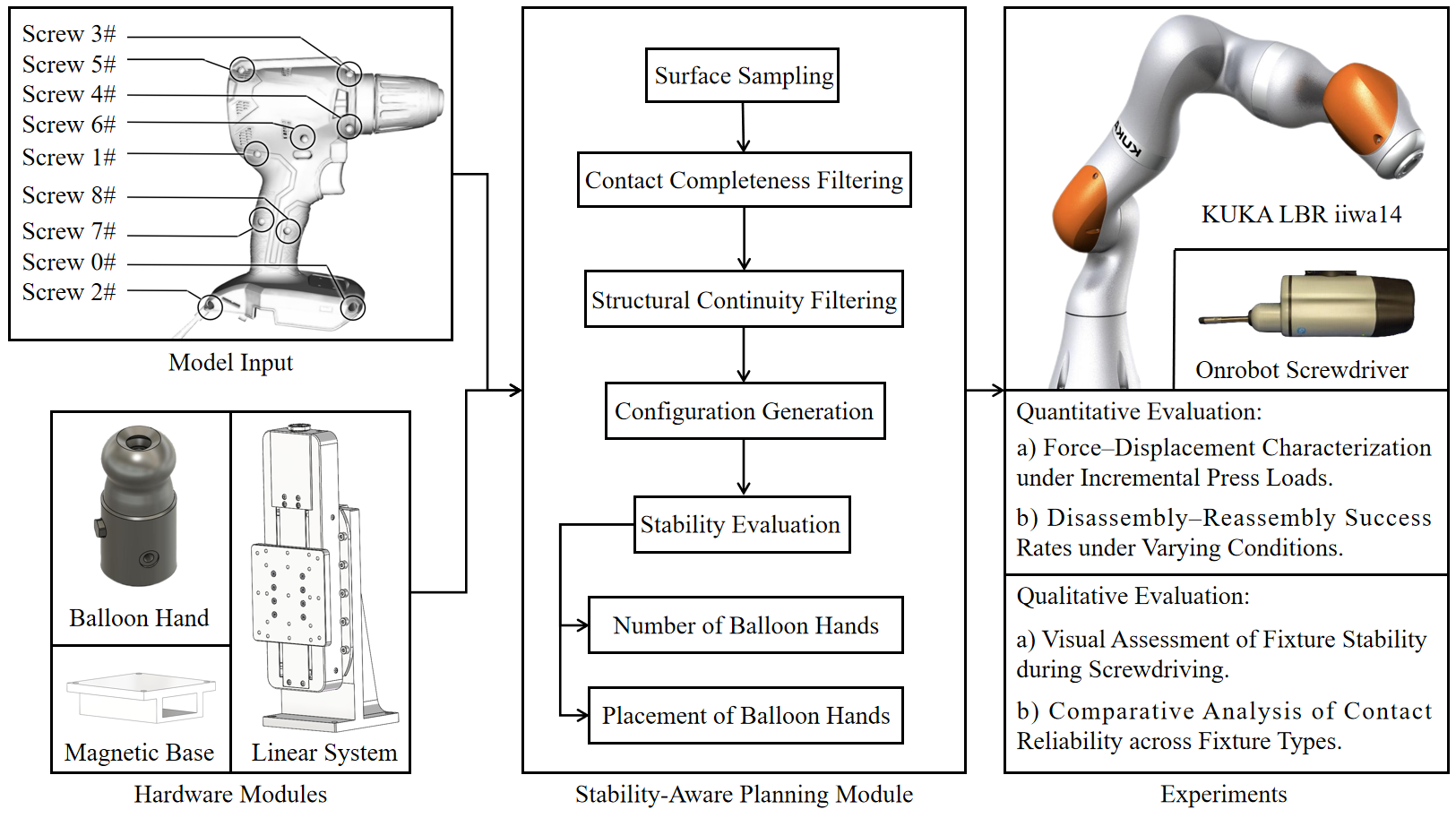}
    \caption{Overview of the system. The input is an object mesh with annotated screw locations. The stability-aware planning module outputs the number and placement of balloon hands. Experiments validate the results using a robotic screwdriver under quantitative and qualitative criteria.}
    \label{fig:system}
\end{figure*}
Screw locations and disassembly sequences are predefined. As illustrated in Fig.~\ref{fig:o}, the experimental system integrates a KUKA LBR iiwa14 robotic arm equipped with an Onrobot electric screwdriver to perform automated disassembly tasks. A balloon-hand-based support mechanism is used to provide adaptive and stable contact with irregular target objects. The support system is mounted on a linear rail and secured via magnetic attachments to a steel surface, ensuring reconfigurable but stable positioning during operation.

\subsection{Modular Vacuum-Based System}

As shown in Fig.~\ref{fig:system}, proposed fixturing system consists of modular vacuum-based end-effectors, each integrating a commercially available balloon-type soft gripper mounted on an adjustable linear actuator. Each linear actuator is mechanically fixed to a magnetic base, which allows secure attachment to a ferromagnetic workspace table. This design enables rapid and flexible deployment without permanent mounting. Each module provides compliant support for irregular object surfaces through localized suction and vertical adjustability. Modules are individually placed according to the contact point configuration computed by the planning algorithm. Although placement is independent, all modules are connected to a shared vacuum source and operate with constant suction pressure.

To balance stability and system simplicity, a minimal-support strategy is adopted, prioritizing configurations that use the fewest number of modules necessary. Preliminary experiments showed that the single-module support was unstable, whereas three-point configurations significantly improved resistance against displacement and rotation during screw disassembly.

\subsection{Stability-Based Planning}
\begin{figure*}[h]
    \centering
    \includegraphics[width=1\linewidth]{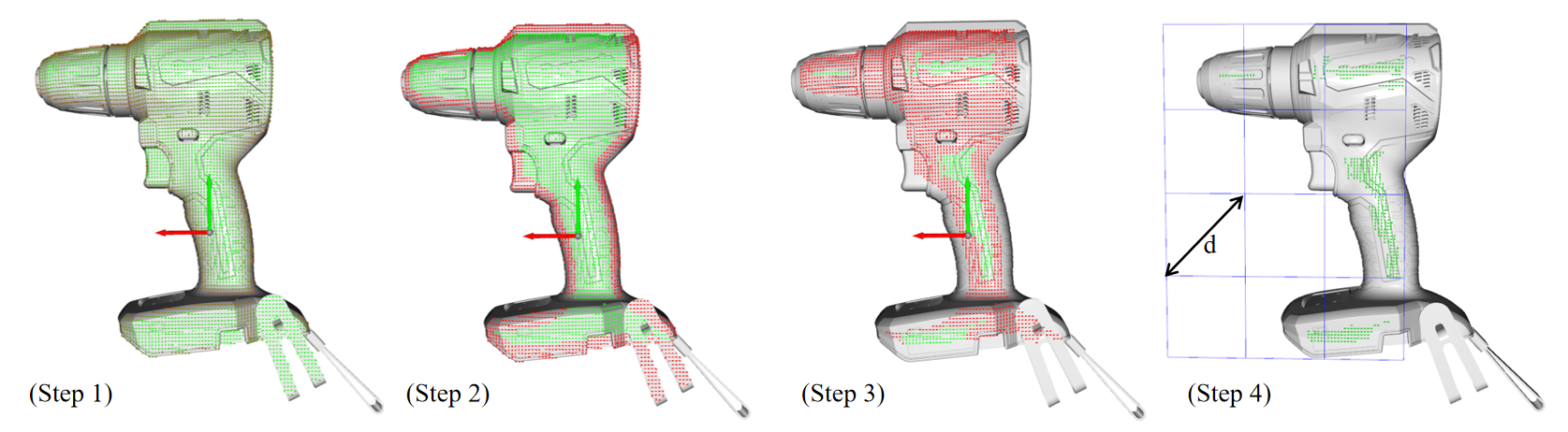}
    \caption{Support point selection pipeline. Step 1: Surface point sampling; Step 2: Contact completeness filtering; Step 3: Structural continuity filtering; Step 4: Configuration generation with spacing constraint ($d = 60\,\mathrm{mm}$).}
    \label{fig:step1-4}
\end{figure*}

To enable a reliable deployment of the proposed soft-fixturing system, a four-stage planning framework shown in Fig.~\ref{fig:step1-4} is introduced to identify feasible support points: surface sampling, contact completeness filtering, structural continuity filtering, and configuration generation. Initial tests showed that single-module support was insufficient, prompting the adoption of multipoint configurations. The three-point contact sets were ultimately chosen based on stability and feasibility evaluations.

For computational efficiency, each balloon hand is approximated as a point contact at its center during planning. The final candidates are then re-evaluated using full-area suction modeling for accurate force estimation and experimental execution.

\textbf{Step 1: Support Point Sampling}

In Step~1 of Fig.~\ref{fig:step1-4}, the planning framework identifies feasible contact regions on the underside of the object for balloon-based fixture modules. Assuming the STL mesh is pre-aligned in a gravity-free pose, screw orientations are considered either horizontal or vertical. In this study, the screws are approximately in the horizontal plane.

Surface sampling is performed by casting vertical rays in the \( +Z \) direction from a 2\,mm-resolution grid over the \( XY \) plane. Each ray records intersections with the mesh, yielding 42{,}495 initial candidate points. Multiple intersections per ray arise because of internal geometry, not just external surfaces.

To refine this raw point set for stable contact analysis, we apply a surface inclination filter to eliminate points located on highly curved regions. For each point \( \mathbf{p}_i \in \mathcal{P}_0 \), we identify its nearest neighbors \( k = 50 \) and perform singular value decomposition (SVD) on the centered local neighborhood. The estimated surface normal \( \mathbf{n}_i \) is taken as the right singular vector corresponding to the smallest singular value. The inclination angle between \( \mathbf{n}_i \) and the upward vertical direction \( \mathbf{z} \) is then calculated as \( \angle(\mathbf{n}_i, \mathbf{z}) \), and the point is retained only if this angle satisfies:
\begin{equation}
\angle(\mathbf{n}_i, \mathbf{z}) \leq \theta, \quad \text{where } \theta = 60^\circ.
\end{equation}
This results in the inclination-filtered set:
\begin{equation}
\mathcal{P}_1 = \left\{ \mathbf{p} \in \mathcal{P}_0 \mid \angle(\mathbf{n}_p, \mathbf{z}) \leq 60^\circ \right\},
\end{equation}
which reduces the initial set to 32{,}633 candidates. The threshold \( \theta = 60^\circ \) is selected to exclude steep or vertical surfaces that are incompatible with suction stability, while retaining gently curved regions suitable for effective contact.

To further refine the set of support candidates, we apply a visibility filter that removes points occluded from the \( -Z \) direction. Although some points in \( \mathcal{P}_1 \) lie on local planar surfaces, they are inaccessible to downward-approaching fixtures due to internal cavities or overhanging structures. We retain only those points that are directly visible from below:
\begin{equation}
\mathcal{P}_2 = \left\{ \mathbf{p} \in \mathcal{P}_1 \mid \mathrm{vis}_{-\mathbf{z}}(\mathbf{p}) = 1 \right\},
\end{equation}

where \( \mathrm{vis}_{-\mathbf{z}}(\mathbf{p}) \) is a binary visibility indicator, equal to 1 if \( \mathbf{p} \) is not obstructed in the \( -Z \) direction.
Next, to ensure gravitational stability, we discard points located above the object's center of mass \( \mathbf{c}_{\text{COM}} \), as such points are prone to tipping and cannot provide stable support:
\begin{equation}
\mathcal{P}_{\text{support}} = \left\{ \mathbf{p} \in \mathcal{P}_2 \mid z_p < z_{\text{COM}} \right\}.
\end{equation}
After all filtering stages, a total of 4{,}099 viable support points remain. These serve as the input to the subsequent contact completeness analysis.

As illustrated in Step~1 of Fig.~\ref{fig:step1-4}, the final filtered point set provides a geometrically and physically plausible foundation for stability-aware fixture planning in the following stages.

\textbf{Step 2: Contact Completeness filtering}

To further ensure that each candidate point can physically accommodate a vacuum-based fixture, we evaluate the contact completeness around each point using a geometric surface coverage criterion derived from the physical constraints of the SGB-10 balloon hand.

For this criterion, we simulate a ring of suction by casting rays vertically upward along a circle of suction radius $r = 8.7\;\mathrm{mm}$, centered at each candidate point $p \in P_{\text{support}}$. Sixty rays are distributed uniformly along the circle in the $XY$ plane. Let $\mathcal{R}(p)$ denote the set of rays at point $p$, and let $\mathcal{H}(p) \subseteq \mathcal{R}(p)$ denote the subset that intersects the mesh within a short vertical distance above the point. A point is considered to have sufficient surface coverage if:

\begin{equation}
\frac{|\mathcal{H}(p)|}{|\mathcal{R}(p)|} \geq \tau,\quad \text{where }\tau = 0.9.
\end{equation}

This condition guarantees that the contact region is not interrupted by large holes or edge transitions that would compromise vacuum sealing. Let the surface-coverage-filtered set be:

\begin{equation}
P_3 = \left\{\, p \in P_{\text{support}} \,\middle|\, \frac{|\mathcal{H}(p)|}{60} \geq 0.9 \,\right\}.
\end{equation}

As shown in Step~2 of Fig.~\ref{fig:step1-4}, 2,753 contact points were retained after all filtering stages. At this stage, we do not yet consider structural support along the Z-axis. Since candidate points in Step 1 were generated from a 2D projection onto the $XY$ plane, any hollowness or internal gaps directly beneath the contact points cannot be determined here. These structural constraints will be addressed in Step 3.

\textbf{Step 3: Structural Continuity Filtering}

\begin{figure}[t]
    \centering
    \includegraphics[width=1\linewidth]{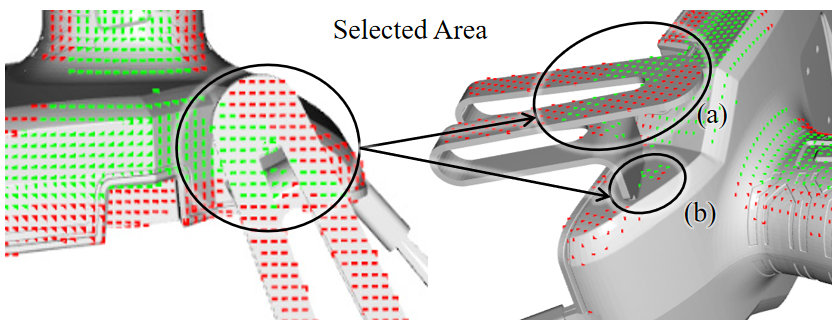}
    \caption{Area (a): Upper part of points in Step~2; Area (b): Lower part of points in Step~2.}
    \label{fig:step2_problem}
\end{figure}

While surface completeness ensures an adequate contact area for sealing, structural continuity beneath each candidate point is also critical to prevent fixture placement in hollow or uneven regions. As illustrated in Fig.~\ref{fig:step2_problem}, many points accepted in Step~2 lie on vertically inconsistent features—e.g., slots, ridges, or embossed patterns—posing risks to suction stability.

To mitigate this, a geometric continuity filter is applied to exclude points on structurally discontinuous surfaces. For each contact point $p \in P_3$, a circular neighborhood $\mathcal{N}(p)$ is defined in the $XY$ plane using the suction radius of the gripper $r = 8.7\,\mathrm{mm}$. The vertical deviation of each neighbor $q \in \mathcal{N}(p)$ is calculated as:
\begin{equation}
    \Delta z_q = |z_q - z_p|.
\end{equation}
Points exceeding a threshold of $\delta = 2.5\,\mathrm{mm}$ are rejected:
\begin{equation}
    P_4 = \left\{\, p \in P_3 \;\middle|\; \forall\, q \in \mathcal{N}(p),\, \Delta z_q \leq \delta \,\right\}.
\end{equation}

As shown in Step~3 of Fig.~\ref{fig:step1-4}, while projection $XY$ suggests local smoothness, discontinuities in $Z$ remain prevalent near cutouts and embossed features. As illustrated in Fig.~\ref{fig:step2_problem}, green points indicate retained candidates; red points violate the continuity condition. This highlights the need to incorporate vertical consistency for robust fixture planning in real-world geometries.

To enable efficient spatial queries, a 2D k-dimensional tree is constructed from the $(x, y)$ coordinates of the filtered candidates. Applying the continuity criterion yields 403 viable support points, which are forwarded to the configuration generation stage.

\section{Evaluations}

This section evaluates the proposed stability-aware planning pipeline via quantitative simulation and hardware validation. Using the filtered candidates from Step~3, valid two-point (2P) and three-point (3P) support configurations are generated under physical constraints of the balloon-based modules.

\subsection{Configuration Generation and Stability Criteria}

As illustrated in Step~4 of Fig.~\ref{fig:step1-4}, the $XY$ plane is partitioned into a uniform grid, with each square cell having a diagonal of $60\,\mathrm{mm}$ to reflect the effective footprint of the balloon hand, thus ensuring the deployability of the fixturing system. This constraint guarantees that at most one candidate support point can be placed in each cell, thereby avoiding collisions between adjacent modules during deployment.

\textbf{Step~4: Configuration Generation} 

From the filtered candidates obtained in Step~4, we generate candidate configurations for both two-point (2P) and three-point (3P) setups. Each configuration is required to satisfy a geometric stability condition based on the object's center of mass $c_{\text{COM}}$. 

To evaluate the geometric stability of each candidate configuration, we adopt a unified convex hull-based inclusion criterion that explicitly accounts for the circular footprint of the suction cups.

Each suction contact point is modeled as a circular region of radius \( r = 8.7\,\mathrm{mm} \). For each point \( p_i \), we generate two auxiliary points by offsetting along the direction orthogonal to the local support plane (typically the XY plane). The set of all offset points is defined as:
\begin{equation}
\mathcal{V} = \bigcup_i \left\{ p_i + r \cdot \mathbf{n}_i,\; p_i - r \cdot \mathbf{n}_i \right\},
\end{equation}
where \( \mathbf{n}_i \) denotes the unit normal vector orthogonal to the baseline at \( p_i \).

This convex-hull inclusion criterion is motivated by static equilibrium considerations. When the object's center of mass lies strictly within the polygon formed by the suction contact region, there exists a set of positive support forces \( \{\mathbf{f}_i\} \) such that both force and moment balance are satisfied under gravity:
\begin{equation}
\sum_{i} \mathbf{f}_i + \mathbf{f}_g = \mathbf{0}, \quad \sum_{i} (\mathbf{p}_i - \mathbf{c}_{\text{COM}}) \times \mathbf{f}_i = \mathbf{0},
\end{equation}
where \( \mathbf{f}_g \) is the gravitational force acting at the center of mass. The existence of such a solution implies that the configuration is capable of resisting tipping or rotation under gravitational loading.

\begin{figure}[t]
    \centering
    \includegraphics[width=1\linewidth]{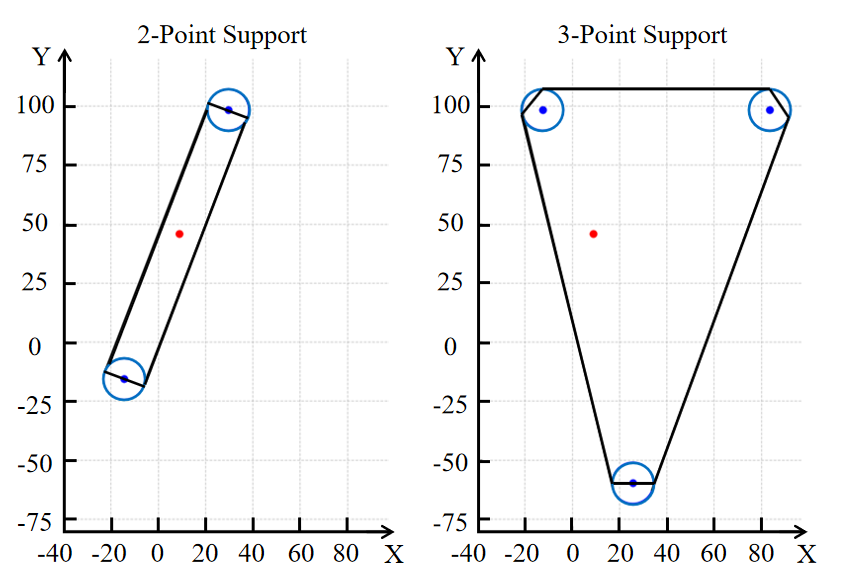}
    \caption{Visualized support configurations on the $XY$ plane. Blue dots indicate the balloon hand contact points; the red dot denotes the object's center of mass.}
    \label{fig:convexhull}
\end{figure}

Therefore, as shown in Fig.~\ref{fig:convexhull}, the inclusion test \( \mathbf{c}_{\text{COM}} \in \mathrm{ConvexHull}(\mathcal{V}) \) serves as a sufficient geometric condition for potential static stability in the 2D projected support plane.

\begin{figure*}[t]
    \centering
    \includegraphics[width=1\linewidth]{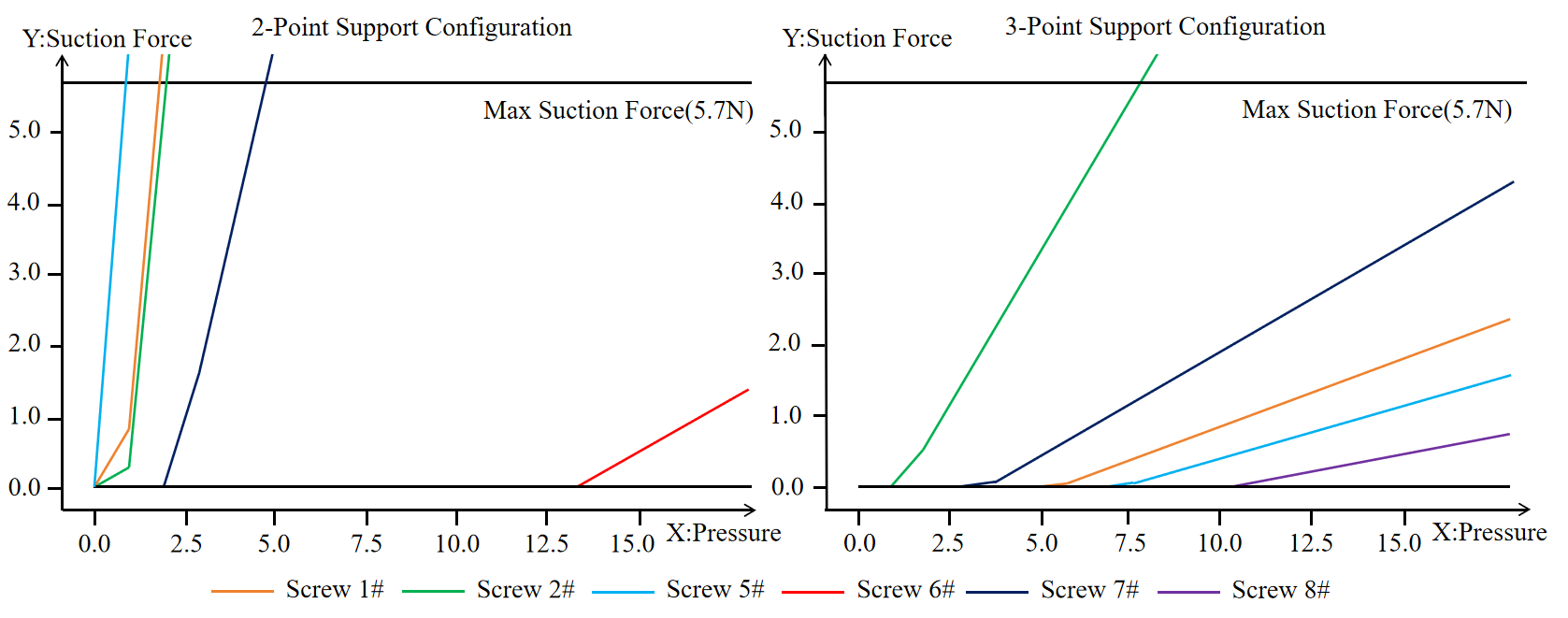}
    \caption{Required suction force per balloon under increasing press force for different screws. 
    The X-axis represents the linearly increased disassembly press force applied by the screwdriver. 
    The Y-axis indicates the corresponding suction force required to maintain equilibrium.}
    \label{fig:ft}
\end{figure*}
\subsection{Stability Analysis of Support Configurations}

To ensure that candidate support configurations can stably hold the object throughout the disassembly process, we evaluated their ability to resist external forces using a static equilibrium framework. Dynamic simulations can, in principle, capture time-dependent effects. They often require detailed modeling of contact mechanics and material parameters, which are difficult to accurately estimate. Therefore, we adopt a quasistatic formulation that emphasizes dominant failure modes such as tipping and slipping while maintaining computational simplicity and robustness.

\subsubsection{Static Equilibrium Model}

We first assess whether each candidate support configuration can maintain static equilibrium when the object is initially placed onto the support structure. The external forces considered include gravity and a constant downward press force applied by the screwdriver upon contact. For each $n$-point support configuration $\{p_j\}_{j=1}^n$, we extract the local surface normals $\{\mathbf{n}_j\}$ from the mesh and define the moment arms $\mathbf{r}_j = p_j - \mathbf{c}_{\text{COM}}$, where $\mathbf{c}_{\text{COM}}$ is the object's center of mass. The static equilibrium system is written as:

\begin{equation}
\underbrace{
\begin{bmatrix}
\mathbf{n}_1^\top \\
\mathbf{n}_2^\top \\
\vdots \\
\mathbf{n}_n^\top \\
(\mathbf{r}_1 \times \mathbf{n}_1)^\top \\
(\mathbf{r}_2 \times \mathbf{n}_2)^\top \\
\vdots \\
(\mathbf{r}_n \times \mathbf{n}_n)^\top
\end{bmatrix}
}_{A \in \mathbb{R}^{6 \times n}}
\cdot
\underbrace{
\begin{bmatrix}
 f_1 \\ f_2 \\ \vdots \\ f_n
\end{bmatrix}
}_{\mathbf{F} \in \mathbb{R}^{n \times 1}}
=
\underbrace{
\begin{bmatrix}
 0 \\ 0 \\ -mg - f_{\text{press}} \\ 0 \\ 0 \\ 0
\end{bmatrix}
}_{\mathbf{b} \in \mathbb{R}^{6 \times 1}}
\end{equation}

The normal contact force computed $f_i$ at each support point is not constrained to be non-negative. If $f_i < 0$, it implies that suction is required to maintain equilibrium. As long as the suction demand satisfies

\begin{equation}
    f_i \geq -f_{\text{max}},
\end{equation}

where $f_{\text{max}} = 5.7\,\mathrm{N}$ denotes the suction limit of a single balloon contact, the configuration remains feasible.

\subsubsection{Quantitative Force Evaluation}
\begin{figure*}[t]
    \centering
    \includegraphics[width=1\linewidth]{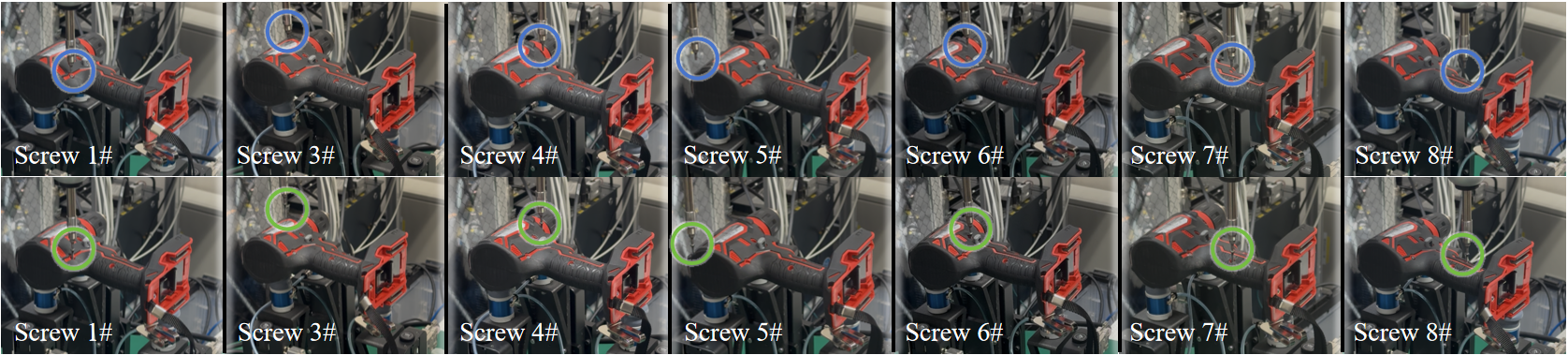}
    \caption{Disassembly of Screws 0\# and 3\#–8\#. Blue: Initial screwdriver position; Green: Successful disassembly.}
    \label{fig:screw}
\end{figure*}
\begin{table}[htbp]
  \centering
  \caption{Pull/Push Forces for Each Screw under 2-Point (2P) and 3-Point (3P) Support}
  \label{tab:pull_push_forces}
  \sffamily
  \resizebox{\linewidth}{!}{%
  \begin{tabular}{c|rr|rrr}
    \toprule
    \multirow{2}{*}{\rule{0pt}{2.5ex}Screw} &
    \multicolumn{2}{c|}{\textbf{2P Configuration} [N]} &
    \multicolumn{3}{c}{\textbf{3P Configuration} [N]} \\
    \cmidrule(r){2-3}\cmidrule(l){4-6}
      & Balloon~1 & Balloon~2 & Balloon~1 & Balloon~2 & Balloon~3 \\
    \midrule
    0 & 262.05 & 280.94 & \phantom{-}2.37 & 21.12 & \phantom{-}6.21 \\
    1 &  -6.10 &  -7.02 & 23.86 &  8.19 &  -2.35 \\
    2 &  -1.50 &  -5.78 & 14.29 & 12.26 &  -6.33 \\
    3 &  35.13 &  67.19 & 14.64 &  1.59 &  13.48 \\
    4 &  86.38 & 115.03 & 11.88 &  6.14 &  11.68 \\
    5 &  -6.80 &  -6.95 & 30.30 &  0.96 &  -1.55 \\
    6 &  -1.51 &   6.50 & 17.65 &  7.00 &   5.06 \\
    7 &  -2.18 &  -6.37 & 19.75 & 14.22 &  -4.27 \\
    8 &  48.05 &  41.69 & 15.42 & 15.02 &  -0.74 \\
    \bottomrule
  \end{tabular}}
\end{table}

As summarized in Table~\ref{tab:pull_push_forces}, the required push and pull forces at each screw location are estimated based on the spatial arrangement of support points. A linearly increasing press force is applied to simulate the disassembly process. The negative force values in the Table~\ref{tab:pull_push_forces} represent the magnitude of the suction force required to maintain equilibrium. When the suction force demand exceeds the maximum capacity of the balloon hand (5.7N), the corresponding screw is deemed unstable.

In the 2-point configuration, screws \#1, \#2, and \#5--\#7 exhibit steep increases in suction demand, surpassing the limit at approximately 2.5N of applied press force, as illustrated in Fig.~\ref{fig:ft}. Screw \#6 remains stable up to 17.5N, indicating enhanced mechanical resilience.

In the 3-point configuration, screws \#1, \#2, \#5, \#7, and \#8 also require significant suction forces. Among these, screw \#2 exceeds the suction limit only beyond 7.5N, reflecting improved force distribution. The others remain within the system's capacity throughout.

Screws \#0 and \#2 are excluded from subsequent real-world success rate analysis due to the unusually deep screw holes, which introduce simulation uncertainty and tool engagement variability.

Although 2-point (2P) support configurations may satisfy static equilibrium in theory, they often fail to provide feasible support in practice due to inadequate moment resistance in 3D space. Screws \#0 and \#4 are notable cases where large reaction forces arise (e.g., 262.05N and 280.94N for screw \#0) when the screw lies far from the line defined by the two contacts. This poor moment geometry leads to excessive force demands beyond the suction capacity, highlighting the insufficiency of 2P setups and motivating the adoption of 3-point (3P) configurations for reliable spatial force closure.

\subsubsection{Force Evolution Under Increasing Press Load}

System behavior during screwdriving is evaluated via a quasi-static simulation, where the applied press force is linearly increased from 0N to 18N. At each increment, static equilibrium is solved and the resulting pull force at each support is examined. Configurations are classified as unstable if any pull force exceeds the suction limit of 5.7N. This procedure offers a tractable means of approximating stability boundaries without resorting to time-dependent modeling.

Force trajectories and final distributions, visualized in Fig.~\ref{fig:ft}, reveal load-sharing patterns and potential risks of overloading individual contacts. The comparison elucidates the mechanical advantages of 3P over 2P support configurations under increasing press loads.

\section{Experiments}

To validate both the feasibility and robustness of our proposed disassembly system, we conducted a series of hardware experiments focusing on real-world screw removal under varying conditions. This section presents multiple experimental evaluations, each designed to test a different aspect of the system performance.

\subsection{Feasibility Evaluation of the Disassembly System}

\begin{figure}[t]
    \centering
    \includegraphics[width=1\linewidth]{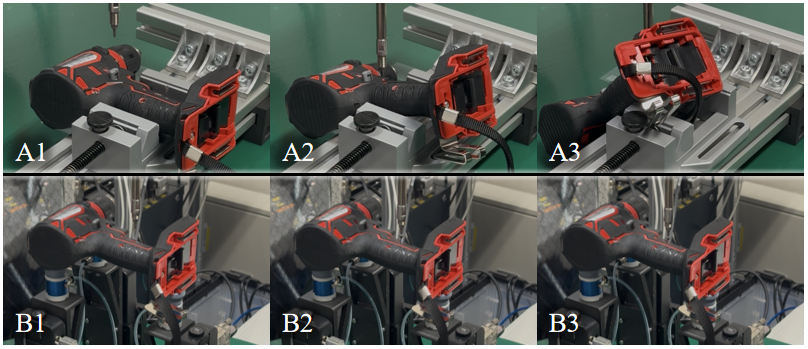}
    \caption{(A1–A3) Disassembly using a traditional rigid gripper. (B1–B3) Disassembly cycle using our proposed system. The screwdriver loosens the screw from B1 to B2 and retightens it from B2 to B3 without re-alignment.}
    \label{fig:vs}
\end{figure}
\begin{figure}[t]
    \centering
    \includegraphics[width=1\linewidth]{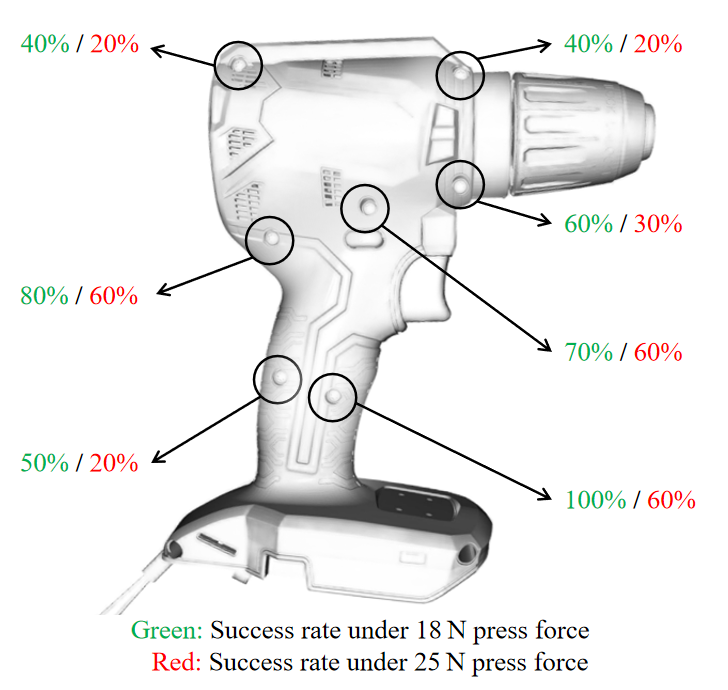}
    \caption{Success rates of disassembly–reassembly cycles for each screw location under two vertical press forces. Green indicates success rate under 18N; red indicates success rate under 25N.}
    \label{fig:sr}
\end{figure}

To verify the feasibility of the proposed disassembly approach, a set of single-pass disassembly trials was conducted across all seven target screws. As shown in Fig.~\ref{fig:screw}, screws \#1 and \#3--\#8 were successfully removed using the integrated system combining balloon-hand support and the OnRobot electric screwdriver. In each trial, the screwdriver was aligned to the target screw (blue), actuated for removal, and the success confirmed upon visual extraction (green). These results demonstrate the practical viability of the support-disassembly integration under realistic conditions.

For comparison, the same task was repeated using a rigid parallel gripper as the supporting structure. As Fig.~\ref{fig:vs} shows, during the disassembly task from A1 to A3, the clamped object tends to gradually tilt or slip due to the predominantly surface or line-based contact between the rigid grippers and the object. Such contact lacks constraint in certain directions, making the system highly sensitive to external disturbances and prone to instability once equilibrium is disrupted. In contrast, the proposed system provides more stable support through localized suction and surface conformity, enabling reliable execution of both screw loosening and tightening operations.

\subsection{Cycle Repeatability Under Varying Press Forces}

A repeated disassembly–reassembly test was performed to assess the stability and robustness of the proposed system. After initial alignment, the screwdriver executed a complete disassembly motion followed by immediate re-tightening of the same screw, without re-alignment between cycles. Each screw underwent up to ten repetitions, and repeatability was measured by the number of successful cycles.

The influence of vertical press force is also examined under two loading conditions: 18N and 25N. For each setting, repeated cycles were executed to evaluate the system’s tolerance to press force variation.

\subsection{Disassembly Success Rate Analysis}

As shown in Fig.~\ref{fig:sr}, under the 18~N press force condition, all screws exhibit an average disassembly success rate above 60\%, whereas higher press forces (25N) lead to noticeable reductions in success rates. This difference arises from two primary factors. First, screws located near the object's center of mass can tolerate larger forces due to improved load distribution. However, increased press force reduces the alignment tolerance between the screwdriver and the screw head. At 18N, the screwdriver can gradually engage with the screw groove even with slight initial misalignment, allowing the system to correct itself during the loosen or tighten phase.

In contrast, excessive press force at 25N may cause the screwdriver to rapidly descend before proper alignment is established, increasing the risk of lateral offset and subsequent failure. Additionally, screws positioned near the edge of the object are more susceptible to misalignment. Due to geometric constraints, any angular deviation may cause the screwdriver to slip along the outer surface rather than entering the screw slot, leading to task failure.

\section{Conclusion and Future Work}

This paper presents a modular, vacuum-based fixturing system designed for stable and repeatable disassembly tasks on irregular objects. A planning framework is introduced to identify optimal support configurations based on geometric filtering and convex hull-based stability analysis. Both simulation and real-world experiments validate the feasibility and robustness of the proposed approach across varying screw locations and press force conditions. The results highlight the advantage of soft, adaptive support in improving success rates and operational repeatability compared to traditional rigid fixtures.

Future work will first explore the automatic selection of balloon hand settings based on the contact area extracted from surface analysis. We also plan to simulate the behavior of the balloon hand under partial deformation, instead of assuming full surface conformity. Furthermore, integrating camera-based recognition during disassembly will help improve the accuracy of contact localization and support adjustment.

\section*{Acknoledgement}
This work was supported by the New Energy and Industrial Technology Development Organization (NEDO) project JPNP23002.

\bibliographystyle{IEEEtran}
\footnotesize
\bibliography{reference}

\end{document}